\documentclass[]{interact}

\usepackage{natbib}
\bibpunct[, ]{(}{)}{;}{a}{}{,}
\renewcommand\bibfont{\fontsize{10}{12}\selectfont}

\usepackage{hyperref}
\usepackage{epstopdf}
\usepackage[caption=false]{subfig}

\theoremstyle{plain}

\theoremstyle{definition}

\theoremstyle{remark}

\begin{document}


\title{A global method to identify trees outside of closed-canopy forests with medium-resolution satellite imagery}

\author{
\name{John Brandt\textsuperscript{a}\thanks{Email: john.brandt@wri.org} and Fred Stolle\textsuperscript{a}}
\affil{\textsuperscript{a}World Resources Institute, 10 G St. NE \#800, Washington D.C., USA}
}

\maketitle

\begin{abstract}
Scattered trees outside of dense, closed-canopy forests are very important for carbon sequestration, supporting livelihoods, maintaining ecosystem integrity, and climate change adaptation and mitigation. In contrast to trees inside of closed-canopy forests, not much is known about the spatial extent and distribution of scattered trees at a global scale. Due to the cost of high-resolution satellite imagery, global monitoring systems rely on medium-resolution satellites to monitor land use and land use change. However, detecting and monitoring scattered trees with an open canopy using medium-resolution satellites is difficult because individual trees often cover a smaller footprint than the satellites’ resolution. Additionally, the variable background land uses and canopy shapes of trees cause a high variability in their spectral signatures. Here we present a globally consistent method to identify trees with canopy diameters greater than three meters with medium-resolution optical and radar imagery. Biweekly cloud-free, pan-sharpened 10 meter Sentinel-2 optical imagery and Sentinel-1 radar imagery are used to train a fully convolutional network, consisting of a convolutional gated recurrent unit layer and a feature pyramid attention layer. Tested across more than 215,000 Sentinel-1 and Sentinel-2 pixels distributed from –60 to +60 latitude, the proposed model exceeds 75\% user’s and producer’s accuracy identifying trees in hectares with a low to medium density ($<40\%$) of tree cover, and 95\% user's and producer's accuracy in hectares with dense ($\geq40\%$) tree cover. In comparison with common remote sensing classification methods, the proposed method increases the accuracy of monitoring tree presence in areas with sparse and scattered tree cover ($<40\%$) by as much as 20\%, and reduces commission and omission error in mountainous and very cloudy regions by nearly half. When applied across large, heterogeneous landscapes, the results demonstrate potential to map trees in high detail and consistent accuracy over diverse landscapes across the globe. This information is important for understanding current land cover and can be used to detect changes in land cover such as agroforestry, buffer zones around biological hotspots, and expansion or encroachment of forests. 
\end{abstract}

\section{Introduction}
\label{}
Forests cover about thirty percent of the world's land surface \citep{fra}. Outside of forests, trees play an important role in agricultural and urban landscapes, as well as in Savannahs, grasslands, and deserts. Although not nearly as dense as forested regions, the vast extent of trees outside of forests contribute greatly to carbon biomass stocks in many countries \citep{Schnell2014TheCO, sequestration}. Trees outside of forests are also important sources of fuel for nearly two-thirds of the world's developing populations \citep{woodfuel}.  However, the spatial distribution of trees outside of closed-canopy forests is not well understood or quantified, despite their importance in carbon sequestration, supporting livelihoods, and the attention given to these trees by major international development agenda \citep{tofreview}. 

The difficulties of identifying trees outside forests with traditional remote sensing methods arise from a combination of the prohibitive expense of analyzing high-resolution imagery at large geographic scales, as well as the sub-pixel sizes of individual trees in medium-resolution imagery. Because of these difficulties, global analyses of dryland forest cover, which tends to be patchy and have an open canopy, have relied on human interpretation of high-resolution satellite imagery rather than remote sensing classifiers \citep{bastin2017}. Within contiguous, closed-canopy forests, applying per-pixel approaches such as bagged decision trees to medium-resolution satellite imagery performs well at monitoring tree cover extent and change due to the high degree of between-pixel similarity in closed-canopy forests \citep{hansen2013, KIM2014178}. However, when considering trees in mosaic landscapes, or trees outside of forests, these medium-resolution, per-pixel approaches are not able to generate reliable maps of tree extent. Even though \citet{radoux2016} found that Sentinel-2, with a ten-meter resolution, has the potential to detect sub-pixel objects as small as three meters, developing globally relevant models that can do so in varying land uses, cloud covers, and terrain has proven very difficult.  
 
Previous approaches to quantitatively map tree cover with remote sensing classifiers have used a variety of supervised and unsupervised machine learning approaches, with either high-resolution or medium-resolution imagery. One of the most extensively used global datasets for forest monitoring is that of \citet{hansen2013}, which developed a global map of tree cover based on Landsat (30 meter resolution) imagery and a random forests classifier. However, because \citet{hansen2013} was designed for monitoring contiguous, closed-canopy forests, it routinely underestimates tree cover in arid landscapes and is not accurate in heterogeneous landscapes such as urban or peri-urban environments \citep{bastin2017, OTTOSEN2020101947}. Indeed, regional models have identified that \citet{hansen2013} underestimates tree cover by up to 80\% in heterogenous landscapes in Europe and underestimates tree cover loss due to small-scale deforestation in the Amazon \citep{OTTOSEN2020101947, Milodowski_2017}. 

At the regional scale, per-pixel machine learning classifiers for medium-resolution satellite imagery (Sentinel-1 and 2) have struggled to detect small patches of trees in heterogeneous landscapes, though these methods are adequate for tracking overall tree cover change at large geographic scales \citep{zhang2019}. Many other recent approaches have used a variety of machine learning methods, such as k-means, random forests, and gradient boosting, to map tree cover at regional scales with Sentinel-2. For instance, \citet{OTTOSEN2020101947} mapped tree cover across Europe with Sentinel-2 and the unsupervised k-means classifier, achieving 53\% user's accuracy and 80\% producer's accuracy on tree detection. Similarly, \citet{Ho_ci_o_2019} developed a regional model of tree species in closed-canopy forests in Poland using Sentinel-2 and the supervised random forests classifier, achieving greater than 80\% overall accuracy distinguishing between eight tree species.  \citet{Immitzer2016FirstEW} mapped tree species in Europe with Sentinel-2 and a supervised random forest model with a 65\% overall accuracy. In a single region in Belgium, \citet{Bolyn2018ForestMA} were able to achieve over 90\% overall accuracy distinguishing forests from non-forests with random forests and Sentinel-2. However, these models focus primarily on homogeneous areas within one geographic region, and do not address the difficulty of sparse tree detection in multiple heterogeneous regions in different biomes.
 
Over the past decade, convolutional neural networks (CNNs) have transformed image processing methods by supplanting both empirically derived classification methods (such as band thresholding or vegetation indices) and per-pixel machine learning methods (such as random forests and support vector machines) with machine learning methods that explicitly learn spatial patterns. Given a set of input images and their labels, a neural network minimizes the differences between the predictions and the labels by learning a set of nonlinear mathematical operations to parameterize the relationship between the input and output domains \citep{goodfellow}. These nonlinear mathematical operations are tuned by optimizing a loss function such as cross entropy or mean squared error with gradient descent \citep{backprop}. CNNs are a special class of neural networks that learn spatial patterns by subsequently applying convolutional operations between the input data and learned matrices of parameters (e.g. weights) \citep{goodfellow}. CNN models have established new state of the art accuracies for a number of remote sensing tasks with high-resolution imagery, such as land-sea segmentation, land use classification, and building identification \citep{Castelluccio2015LandUC, Li2017DeepUNetAD, ulu2018, ZHANG201857}.
 
Although deep learning methods have come to popularity in recent years, their applications largely remain limited to high-resolution (less than 3 meter) imagery, which are often too expensive or computationally intense for large area analyses in real-world applications. In their review of deep learning approaches in remote sensing, \citet{ma2019} identified five barriers to applying deep learning to medium-resolution satellite imagery. These include a tendency of papers to focus on small geographic regions, the lack of fine structural details in medium-resolution imagery, difficulties with differently scaled objects, a lack of studies using time series imagery, and the tendency for deep learning approaches to generate blurry outputs. In homogeneous and geographically small landscapes, such as monoculture plantations, deep learning and medium-resolution sentinel imagery can classify tree density with higher than 80\% accuracy \citep{countinguncountable}. However, existing deep learning models struggle to generalize to new geographies because background land uses can contribute more to the spectral information of the corresponding pixel than the tree itself due to the relative size differences between trees and satellite pixels. This noisiness is further complicated by the observation that many popular deep learning models generate fuzzy outputs with degraded accuracy around boundaries of objects \citep{IBTEHAZ202074, pyramid}. While this may be sufficient for high-resolution imagery, small patches of trees in medium-resolution imagery almost entirely consist of boundary pixels and thus classifiers must be designed with this constraint in mind.

While most CNN approaches rely on a single image input, satellite systems are designed to capture sometimes dozens of images per year of each location. Time series convolutional neural networks analyze a temporal sequence of imagery rather than a single input image. The temporal domain adds additional complexity to the satellite imagery that is especially useful for low- and medium-resolution satellite imagery where the complexity of each individual image alone may not be sufficient for deep learning. Additionally, time series models benefit from limiting the noise in individual satellite images driven by atmospheric conditions and the relative satellite and sun positions. Time series models, such as the convolutional long-short-term memory (cLSTM) and convolutional gated recurrent unit (cGRU) learn spatio-temporal relationships by extending the convolution operation to the temporal dimension \citep{convlstm}. Each time step image is convolved with the prior time step and weight matrices that decide how much of the short term information to keep (e.g. forget gate) and how much long-term information to keep (e.g. reset gate). This allows for the generation of complex models of both short-term changes, such as leaf-out events, and long-term seasonal patterns. Indeed, the accuracy of medium-resolution remote sensing models have recently seen improvement by using multi-temporal image analysis \citep{rubworm2018,Roy2019MultitemporalLU}. Another recent advancement in medium-resolution remote sensing classification is the fusion of Sentinel-1 radar data with Sentinel-2 optical data, which has been shown to increase the accuracy of land use and land cover mapping in cloudy regions \citep{doi:10.1080/22797254.2019.1596757, tavares2019}.

Recent developments in the field of computer vision include many modifications to CNN models that bring important improvements to the generalizability, pixel-level boundary accuracy, and performance on unbalanced classes, though their applicability to medium-resolution remote sensing models have often not been established. With regard to generalizability, new normalization methods such as layer normalization and batch renormalization standardize intermediate layers to add stochasticity and reduce the extreme predictions on test data that are outside of the range of the training data \citep{batchrenorm, layernorm}. Squeeze and excitation layers force CNNs to learn finely-grained filters which reduce blurriness by rescaling the outputs based on a learned scoring map for either channels or pixels \citep{csse}. New loss functions, such as the focal loss, which weighs hard-to-classify samples more, the Lov{\'{a}}sz-Softmax loss, which focuses on regions rather than pixels, and the boundary loss, which focus on boundaries between classes, have greatly increased the abilities of CNN models to perform well in highly unbalanced classification scenarios, such as medical imagery analysis and land use classification \citep{focalloss, boundaryloss, DBLP:journals/corr/BermanB17}. Extending on these approaches, this paper uses fused multi-temporal imagery from Sentinel-1 and Sentinel-2 to construct a deep learning model that generates robust classifications of tree presence at the ten meter scale across a variety of geographies, terrains, and land uses.   
 
\section{Materials and Methods}
\subsection{Data}

\begin{figure}
\begin{center}
\begin{tabular}{c}
\includegraphics[height=4cm]{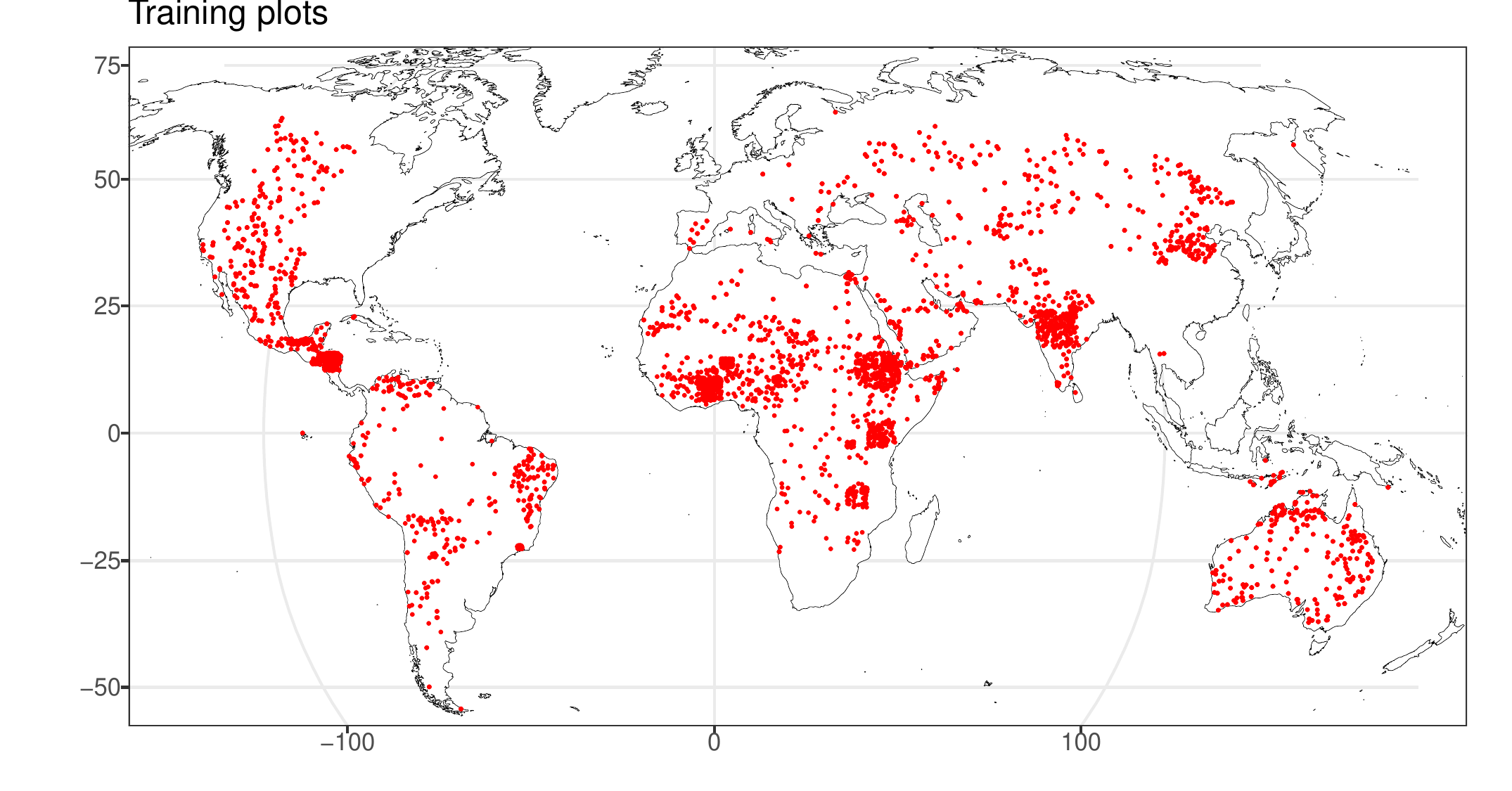}  
\end{tabular}
\end{center}
\caption 
{ \label{fig:training_plots}
Locations of training sample plots. Each sample plot is 140 x 140 meters (approximately 2 hectares) with a 10 meter pixel size.} 
\end{figure}

\begin{figure}
\begin{center}
\begin{tabular}{c}
\includegraphics[height=4cm]{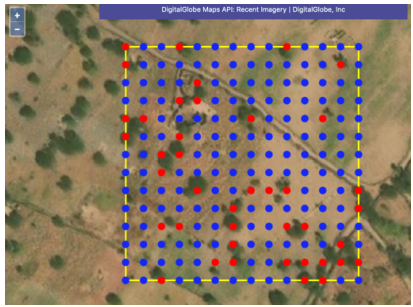}  
\end{tabular}
\end{center}
\caption 
{ \label{fig:plot_labeling}
Screenshot from Collect Earth Online showing the plot labelling process. Red grid cells indicate the presence of a tree, while blue grid cells indicate the absence.}
\end{figure}

A total of 4,500 training sample plots, distributed semi-randomly from -60 to +60 latitude, were labelled with visual interpretation of high-resolution imagery on Collect Earth Online, an online platform for systematically labelling geospatial data with high-resolution imagery (0.5 meters, WorldView 3, Figure \ref{fig:training_plots}) \citep{collectearthonline}. Sample plots were sized 140 x 140 meters with sampling points positioned within at 10 meter intervals for 196 samples per plot. Pixels were marked positive if they intersected a tree identified through visual interpretation of a cloud-free, leaf-on high-resolution image (Figure \ref{fig:plot_labeling}). The presence of a tree was determined based on the surrounding land use, presence of a shadow, and size relative to identifiable shrubs or grass in proximity. Trees with a canopy diameter smaller than three meters were excluded from consideration. Because tree presence was mapped at the 10 meter scale, the presence of multiple trees within each 10 meter pixel was not differentiated from the presence of a single tree.

Tree presence at the ten meter scale was predicted with fused Sentinel-1 imagery, Sentinel-2 imagery, and slope derived from the MapZen digital elevation model (DEM). Sentinel-2 detects 13 bands with 10, 20 and 60 meter resolution, including the visible, near infrared, and short-wave infrared spectrums. Sentinel-1 provides 10 meter synthetic aperture radar (SAR) data of the entire world every 12 days. Twenty-four cloud-free images of each 19,600 m\textsuperscript{2} (140 x 140 m) sample area, separated by fifteen days each, were created by removing and interpolating cloud cover and shadow from each Sentinel-2 image acquisition (process described below) and fusing the nearest Sentinel-1 image from January 1 to December 31 2019. For Sentinel-2, the 10 and 20 meter bottom-of-atmosphere (L2A) bands were selected. For Sentinel-1, VV-VH imagery with the gamma back scatter coefficient was used. Data was accessed through the Sentinelhub API. Twenty-meter Sentinel-2 bands were upscaled to ten meters with DSen2, which is a convolutional neural network (CNN) approach to pan-sharpening Sentinel-2 imagery \citep{dsen2}. Clouds were identified with \href{https://github.com/sentinel-hub/sentinel2-cloud-detector}{S2Cloudless} and cloud shadows were identified by generating a mask with the methodology proposed in \citet{candra2020} and removing cloud shadow masks which were more than 800 meters from an identified cloud. Images with more than 25\% cloud or shadow cover were removed, and remaining clouds and cloud shadows greater than 250 m\textsuperscript{2} (25 pixels) were linearly interpolated with pixels from the nearest two clean time steps after which the Whittaker smoother was used ($\lambda$ = 800, d = 2) to interpolate missing pixels \citep{whittaker}. 
 
The DEM was degraded with a 5x5-pixel median filter before calculating slope to reduce noise. In addition to the raw band values, the enhanced vegetation index (EVI, Eq. (1)), modified soil adjusted vegetation index (MSAVI2, Eq. (2)), and bare soil index (BI, Eq. (3)) were calculated and included as model input \citep{evi, msavi, baresoilindex}. Additional indices, including normalized difference vegetation index, soil adjusted vegetation index, and normalized difference moisture index, were tested but did not improve performance. 

\begin{equation} \label{evi-eq}
\textrm{EVI} = 2.5\dfrac{(B8 - B4)}{(B5 + 6 * B4 - 7.5 * B2 + 1)}
\end{equation}

\begin{equation} \label{msavi-eq}
\textrm{MSAVI2} = \dfrac{2 * B8 + 1-\sqrt{(2*B8 + 1)^2 - 8(B8-B4)}}{2}
\end{equation}

\begin{equation} \label{bi-eq}
\textrm{BI} = \dfrac{B2 + B4 - B3}{B2 + B4 + B3} 
\end{equation}

\subsection{Model}
\begin{figure}[!h]
\begin{center}
\begin{tabular}{c}
\includegraphics[height=3.6cm]{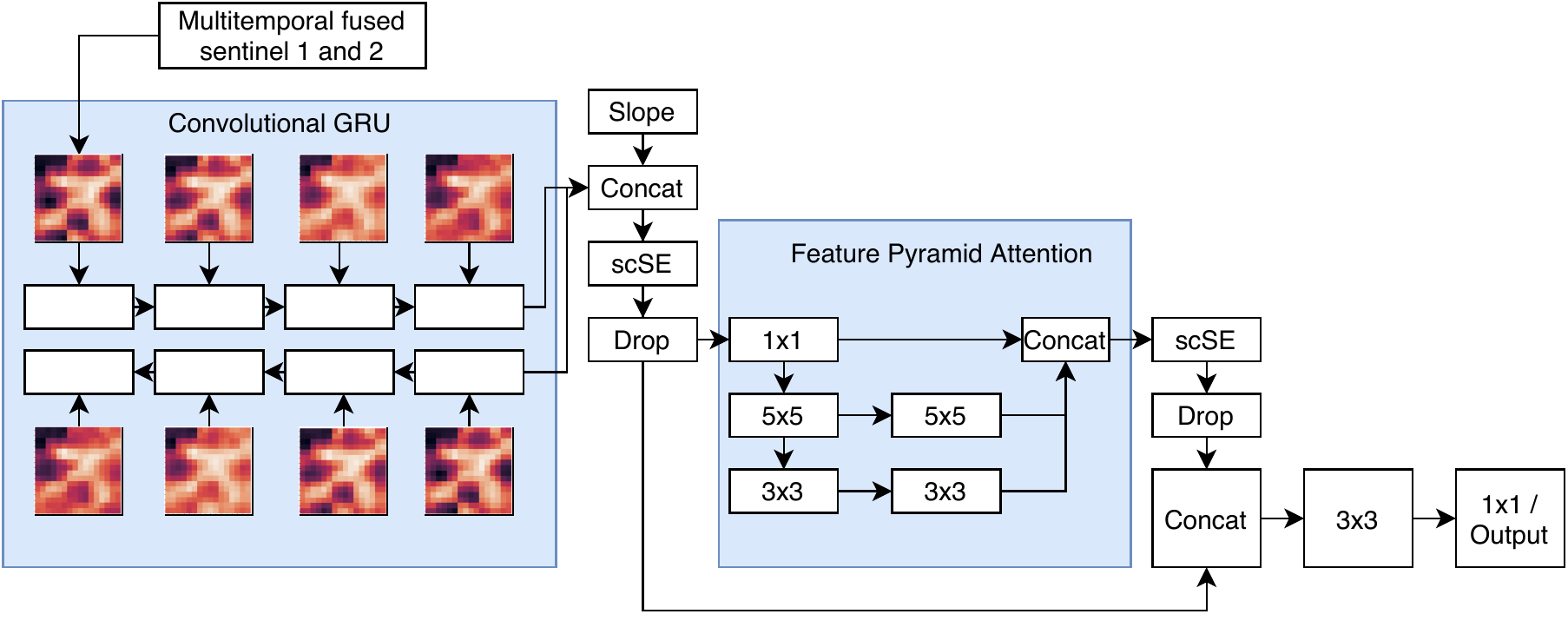}  
\end{tabular}
\end{center}
\caption 
{\label{fig:network}
Overview of network architecture. Twenty-four input images for ten Sentinel-2, three indices, and two Sentinel-1 bands are passed to a bidirectional convolutional GRU to model temporal relationships. Next, the feature pyramid attention module integrates regional information with pixel-level information. Finally, Conv-BN-RELU-csSE-DropBlock blocks are combined with hypercolumns before sigmoid classification.} 
\end{figure}

The model consists of a fully convolutional neural network with a bidirectional convolutional gated recurrent unit (cGRU) encoder and a feature pyramid attention (FPA) decoder \citep{convlstm, pyramid} (Figure \ref{fig:network}). The cGRU takes as input the biweekly processed Sentinel-1 and Sentinel-2 bands, and uses two-dimensional 3 x 3 convolutions to generate feature encodings that represent the per-pixel change over time. The FPA layer takes the local features generated by the cGRU and increases their field of view, incorporating knowledge about features from up to 15 pixels away, while maintaining the fine-grained localization of feature maps. This is done by multiplying deeper convolutional layers by a 1x1 convolutional layer. The upsampling blocks in the FPA layer use an upsize convolution rather than a transpose convolution in accordance with the recommendations in \citet{odena2016deconvolution}. These feature maps, which incorporate knowledge of both per-pixel and regional change over time, are then classified with a standard convolutional layer with a sigmoid activation.  

The cGRU layer uses layer normalization which standardizes features to avoid exploding gradients, and a channel squeeze and spatial excitation (SE) layer, which improves fine grained localization, between time steps \citep{layernorm, csse}. The SE block rescales each input channel c by multiplying by a sigmoid activated 1 x 1 x c convolution. Separate SE blocks are learned for the update and the reset gate of the GRU, respectively, and weights are shared between time steps. 
 
The FPA layer is followed by two convolutional blocks with batch renormalization and csSE (Fig. 6) \citep{batchrenorm, csse}. Hypercolumns, which concatenate deep and shallow features to increase layer dimension, are placed before the final sigmoid classification layer to facilitate pixel-level accuracy \citep{hypercolumns}. The sigmoid bias layer was initialized in the same manner as in \citet{focalloss} to avoid model collapse in the early stages of training. All padding operations in the model used reflect padding to enforce the distributional consistency of border convolutions. Weights were initialized according to \citet{heinitializer} when combined with batch renormalization, and glorot uniform otherwise \citep{glorotuniform}. Non-linear activations for intermediate layers use the rectified linear unit (RELU) \citep{Nair2010RectifiedLU}. The ‘rmax’ and ‘dmax’ batch renormalization parameters follow the learning schedule in the original paper \citep{batchrenorm}. All experiments were conducted in Tensorflow 1.13.1 and the presented model has 221 thousand learnable parameters \citep{tensorflow2015-whitepaper}. Data and code are released for reproducibility on the \href{https://github.com/wri/restoration-mapper}{author's GitHub page}. 
 
The model was optimized with the Adabound optimizer with learning rates between 1e-4 and 2e-2, and was trained for 100 epochs on a NVidia K80 GPU \citep{adabound}. A batch size of 20 was selected with equibatch sampling by tree cover percent \citep{DBLP:journals/corr/BermanB17}. To mitigate overfitting, several regularization methods were used. These include zoneout (prob = 0.2) in the GRU layer and dropblock (prob increases from 0 to 0.2 during training) after each intermediate convolutional layer \citep{zoneout, dropblock}. The loss function for the model was a combination of label-smoothed (prob = 0.10) binary cross entropy, which was weighted by the effective number of samples, and the boundary loss \citep{labelsmooth, weighting, boundaryloss}. The weight of the boundary loss (BL) increased proportionate to the cross entropy loss (CE) throughout training according to Equation \eqref{eq1}, where $y$ is the label segmentation, $p$ are the sigmoid probabilities, $\Omega$ is the spatial domain of $y$, and $\phi_G$ is a distance map with respect to the boundary of the positive segments of $y$.

\begin{equation} \label{eq1}
\begin{split}
\textrm{Loss} &= (1-\alpha) * \textrm{CE} + \alpha* \textrm{BL} \\
\alpha &= \textrm{Epoch} * 0.01, \alpha\in[0.0, 0.5] \\
\textrm{BL} &= \int_{\Omega} \phi_G(y)p_\theta(y)dy \\
\textrm{CE} &= - (y \textrm{log}(p\in[0.1, 0.9]) + (1-y)\textrm{log}(1-p\in[0.1, 0.9])) \\
\end{split}
\end{equation}

The model outputs were post-processed to improve accuracy around the border of input tiles by smoothing the predictions over a tiled window with two-dimensional interpolation between overlapping patches. The prediction for each pixel was calculated as the weighted average of nine input images, which were shifted either up, left, down, right, or diagonally by 70 meters from the original bounds, weighted by the distance of each pixel to the center of each of the nine input images with a Gaussian filter with a 3.5 pixel standard deviation. Prediction probabilities were converted to binary class labels based on a threshold determined by the receiver operating characteristic.

\subsection{Validation Methods}
Model performance was assessed at the global scale for pixel-level accuracy of tree identification and plot-level (140 x 140 meter) tree cover accuracy. Tree cover was calculated for each plot as the proportion of 10 x 10 meter pixels which were marked positive for tree presence in the same manner as in \citet{bastin2017}. Model performance was assessed for each decile of plot-level tree cover to identify comparative performance across different tree densities and distributions.  In addition to the global validation methods, model performance was also tested for pixel and tree cover accuracy in three selected one million hectare landscapes in different biomes in Tanzania, Ghana, and Honduras. In all cases, accuracy validation was assessed against human-annotated high-resolution imagery, a method which has previously been used for accuracy assessments of wall-to-wall maps \citep{hansen2013, zhang2019}.

The proposed modelling approach was also tested against standard baselines, including random forests (RF), support vector machines (SVM), and a U-net CNN \citep{unet}. The U-net is an often used CNN approach for remote sensing classification \citep{DBLP:journals/corr/abs-1904-00592, Li2017DeepUNetAD}. Baseline models were tested with different temporal aggregation methods, including mean, median, standard deviation, and quarterly and monthly mean composites. Both the random forests and the support vector machine baseline models use the mean band reflectance for each pixel over the 24 cloud-free images because this approach outperformed the other aggregation methods. The U-net baseline uses the median band reflectance. Common hyperparameters for each baseline approach were tuned using a brute force approach. The U-net CNN was designed in a similar manner to \citet{unet}, with the additions of reflect padding, batch renormalization, dropblock, and upsample convolution to match design choices of the proposed model. The U-net has 620 thousand trainable parameters and uses the Adabound optimizer with learning rates between 0.001 and 0.1.

\subsubsection{Metrics}

\begin{figure}
\begin{center}
\begin{tabular}{c}
\includegraphics[height=6cm]{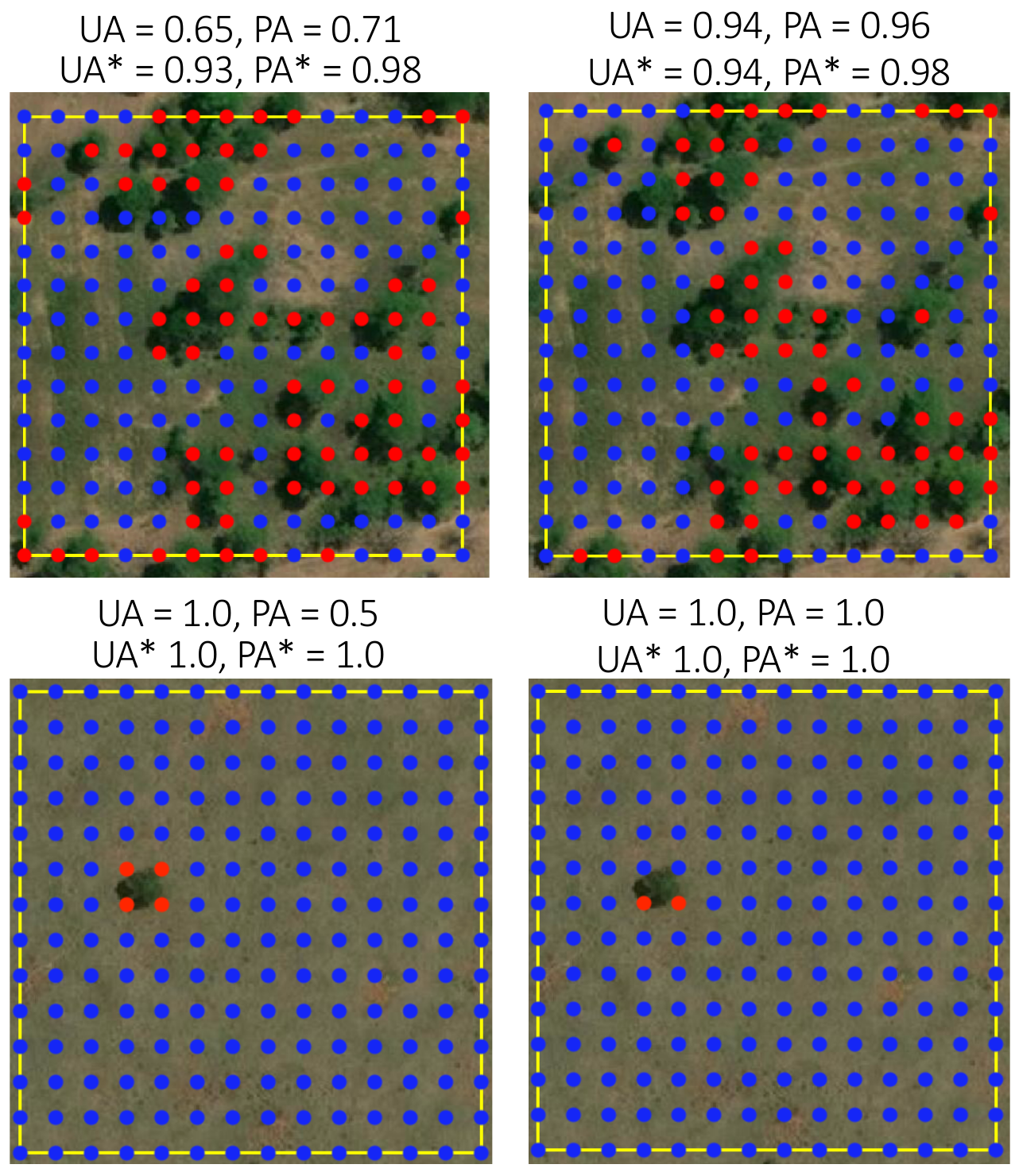}  
\end{tabular}
\end{center}
\caption 
{ \label{fig:coregistration}
The user's (UA) and producer's accuracy (PA) are sensitive to labelling choices due to coregistration errors between Sentinel and WorldView imagery. Red pixels denote trees, blue pixels denote background. The labelling scheme on the left results in low UA and PA scores when analyzed with Sentinel imagery, despite being visually accurate. The labelling scheme on the right achieves high UA and PA scores. Both labelling schemes achieve similar UA and PA scores (denoted with *) when adjusted with Equation \eqref{eq5}.} 
\end{figure}

Model performance was evaluated with the user's and producer's accuracy. These metrics were modified in order to mitigate errors caused by variable coregistration consistency between WorldView 3 and Sentinel-2 imagery. Sub-pixel shifts in the Sentinel imagery are caused by resampling each image with nearest neighbor interpolation to fit within the plot boundaries due to the average coregistration error between images of 12 meters \citep{sentinelqc}. Additionally, because WorldView 3 and Sentinel-2 have different viewing geometries and use different digital elevation models (DEM) for orthorectification, the products may be locally misaligned by more than ten meters \citep{K_b_2016, worldviewfuse, sentinelworldviewfuse}. These local misalignments cause shifts between the WorldView 3 labels and the Sentinel predictions, rendering per-pixel metrics useless in many cases (Figure \ref{fig:coregistration}). To account for this, the producer's accuracy counts false negatives (FN) if the ground truth positive occurs more than ten meters from a predicted positive at location $(x, y)$. Similarly, the user's accuracy counts false positives (FP) if the predicted positive occurs more than ten meters from a positive ground truth pixel. There was no double counting of trees in either the ground truth or the predictions. The metrics are calculated according to Equation \eqref{eq5}, where TP refers to true positives, $\hat{Y}$ refers to the prediction, and $Y$ refers to the label.

\begin{equation} \label{eq5}
\begin{split}
\textrm{User's acc} &= \dfrac{\textrm{TP}}{\textrm{TP + FP}} \\
\textrm{Producer's acc} &= \dfrac{\textrm{TP}}{\textrm{TP + FN}} \\
\textrm{TP} &= \sum_{x,y} \big((Y_{x,y}) (\max\hat{Y}_{x-1:x+1,y-1:y+1})\big) \\
\textrm{FP} &= \sum_{x,y} \big((\hat{Y}_{x,y}) (1 -\max Y_{x-1:x+1,y-1:y+1})\big) \\
\textrm{FN} &= \sum_{x,y} \big((Y_{x,y}) (1 - \max \hat{Y}_{x-1:x+1,y-1:y+1})\big)
\end{split}
\end{equation}

\subsubsection{Global validation}
\begin{figure}[!h]
\begin{center}
\begin{tabular}{c}
\includegraphics[height=5cm]{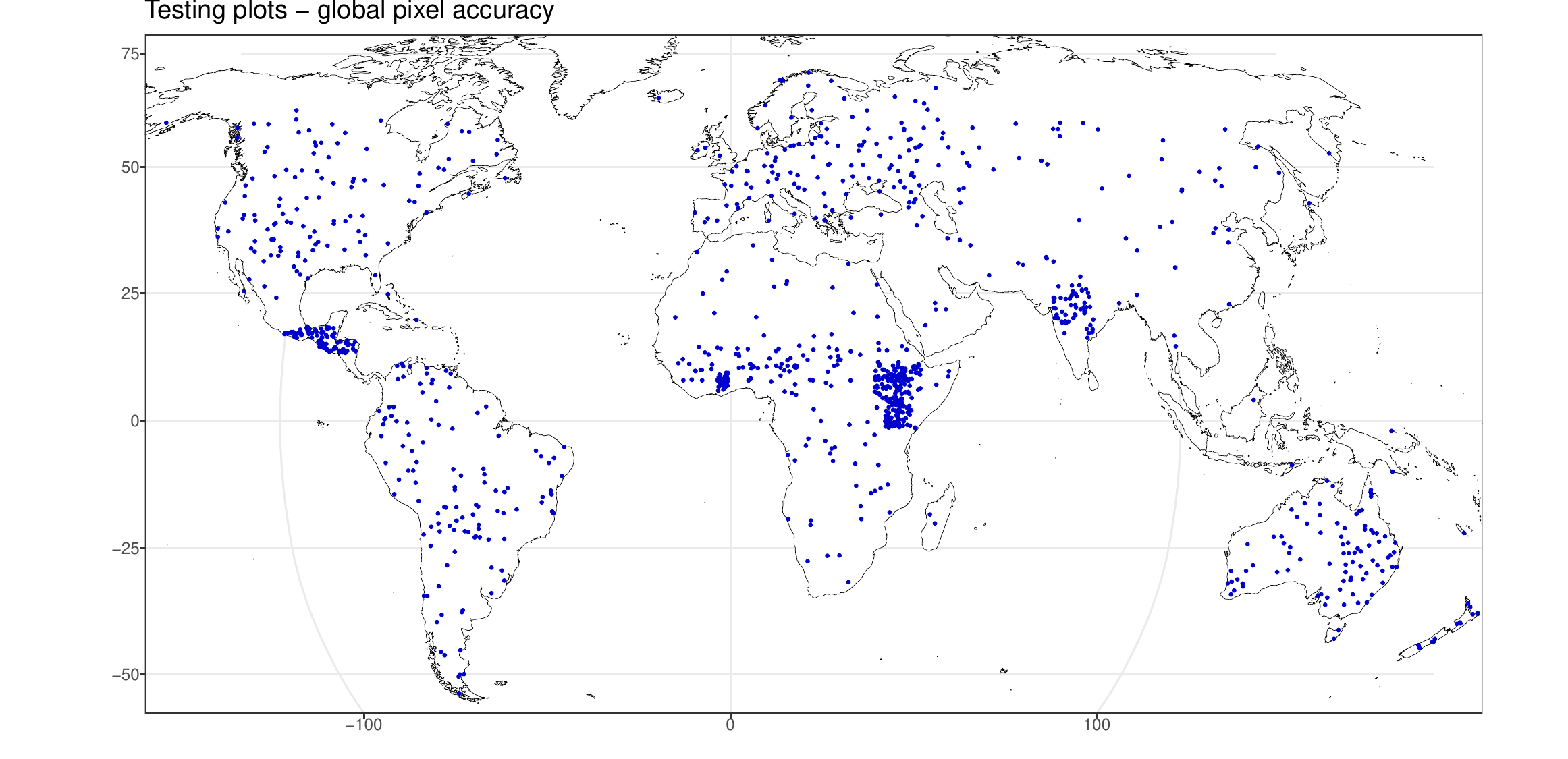}  
\end{tabular}
\end{center}
\caption 
{ \label{fig:test_plots}
Locations of testing sample plots. Each sample plot is 140 x 140 meters (approximately 2 hectares) with a 10 meter pixel size.} 
\end{figure}

Pixel-level performance was assessed with 1,100 140 x 140 meter plots labelled in the same manner as the training data (section 2.1) (Figure \ref{fig:test_plots}). Seven hundred of the test plots were randomly distributed on land areas between -60 and +60 latitude. The other four hundred were randomly distributed within selected regions (including Central America, West Africa, East Africa, and India) to increase the ratio of test plots with high cloud cover and difficult land uses (such as plantation agriculture, agroforestry, step agriculture, and urbanization).

\begin{figure*}[!h]
\begin{center}
\begin{tabular}{c}
\includegraphics[height=5.25cm]{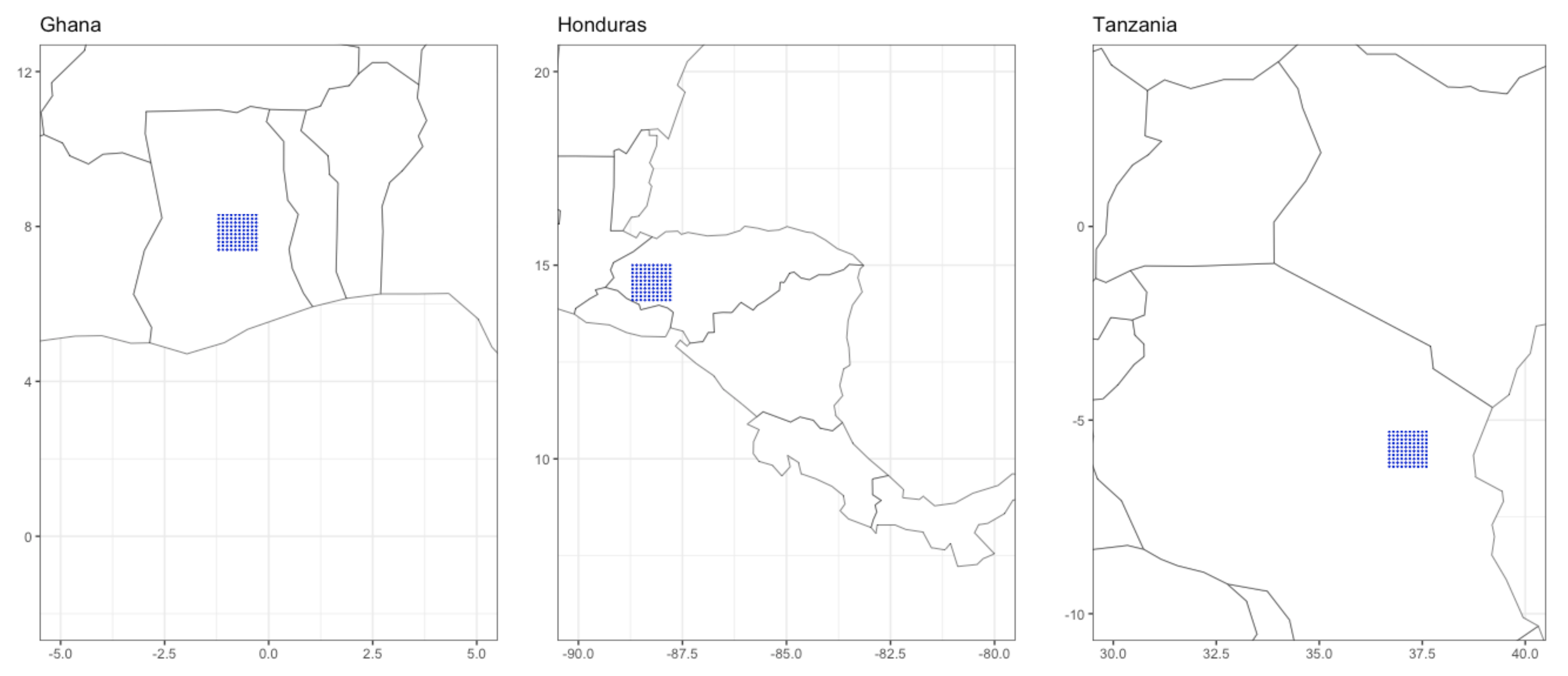}  
\end{tabular}
\end{center}
\caption 
{ \label{fig:region_plots}
Locations of regional sample plots. Within each of the three regions, 100 evenly distributed 140 x 140 meter plots (represented with blue dots) were labelled with a 10 meter pixel size for binary tree presence. Additionally, within each of the three regions, a randomly selected 200 hectare sub-region was labelled with a 10 meter pixel size for binary tree presence.} 
\end{figure*}

\begin{figure}[!htbp]
\begin{center}
\begin{tabular}{c}
\includegraphics[height=5cm]{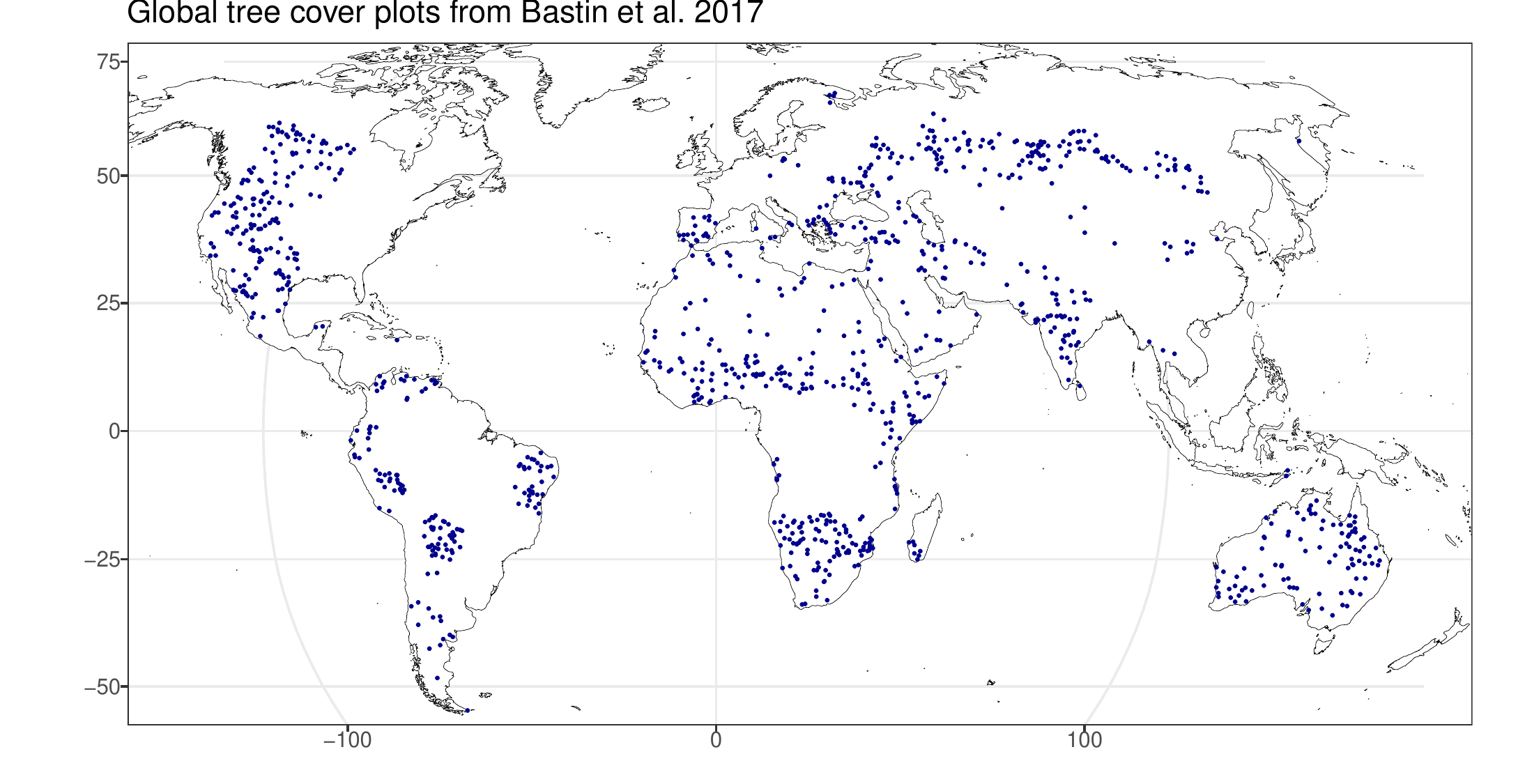}  
\end{tabular}
\end{center}
\caption 
{ \label{fig:bastin-locations}
Locations of 1,000 half-hectare plots from \citet{bastin2017} stratified by tree cover decile.}
\end{figure}

To further assess model performance at the global scale, the percent tree cover of a random sub-sample of 1,000 half-hectare plots across global drylands from \citet{bastin2017}, stratified by geography and tree cover decile, were reanalyzed with recent high-resolution imagery (Figure \ref{fig:bastin-locations}). The percent tree cover was labeled in the same way as in \citet{bastin2017}, as the proportion of 10 meter pixels per plot that intersect a tree canopy. Thirty five percent of the reanalyzed plots disagreed with the labels in \citet{bastin2017} by more than 30\% tree cover. The labels for these plots were reassigned based on recent high-resolution imagery, while the rest retained their original labels. One source of disagreement was due to land use change since the original analysis. Because the plots were stratified by tree cover percent, plots with partial tree cover, were over-represented. These plots are also more likely to experience land use change than are plots with barren or dense canopy \citep{woodyencroachment, kalamandeen}. Another source of label disagreement arose from coregistration errors. Because more than half of 10-meter pixels in a half-hectare plot are border pixels, coregistration errors under 10 meters between the image analyzed in \citet{bastin2017} and recent high-resolution imagery can alter tree cover predictions by up to 50\%. Finally, \citet{commentbastin} identified significant classification error in the \citet{bastin2017} data set due to old imagery and a lack of quality assurance. Tree density was calculated as the proportion of plot pixels with predictions higher than a threshold determined by the receiver operating characteristic. 

\subsubsection{Regional validation}

Model performance was also tested to ensure that the results for large geographic regions were consistent across different land uses and geographies. To do this, we prepared wall-to-wall maps of one million hectare regions in three geographies (Figure \ref{fig:region_plots}). The selected geographies included a Semi-arid desert with agricultural production in Tanzania, a tree Savannah with agricultural production in Ghana, and a mosaic tropical broad leaf forest landscape in Honduras. Model performance was assessed for these three regions through two complementary methods. To assess performance over the total one million  hectares, 100 evenly distributed 140 x 140 meter plots were labelled in the same manner as in section 2.1. Additionally, to assess performance across smaller regions, a 200 hectare subregion was randomly identified for each of the three regions, for which each 10 meter pixel was labelled for binary tree presence. Combining these two validation sources, a total of 39,600 pixels were evaluated for each of the three regions.  We also compare predictions with those of \citet{zhang2019}, who used a support vector machine to identify trees in 3 years of fused Sentinel-1 and Sentinel-2 imagery, with training data consisting of 20,000 labelled pixels in high-resolution imagery across the western Sahel. 

\section{Results}

\subsection{Global accuracy}

\begin{figure*}
\begin{center}
\begin{tabular}{c}
\includegraphics[height=6cm]{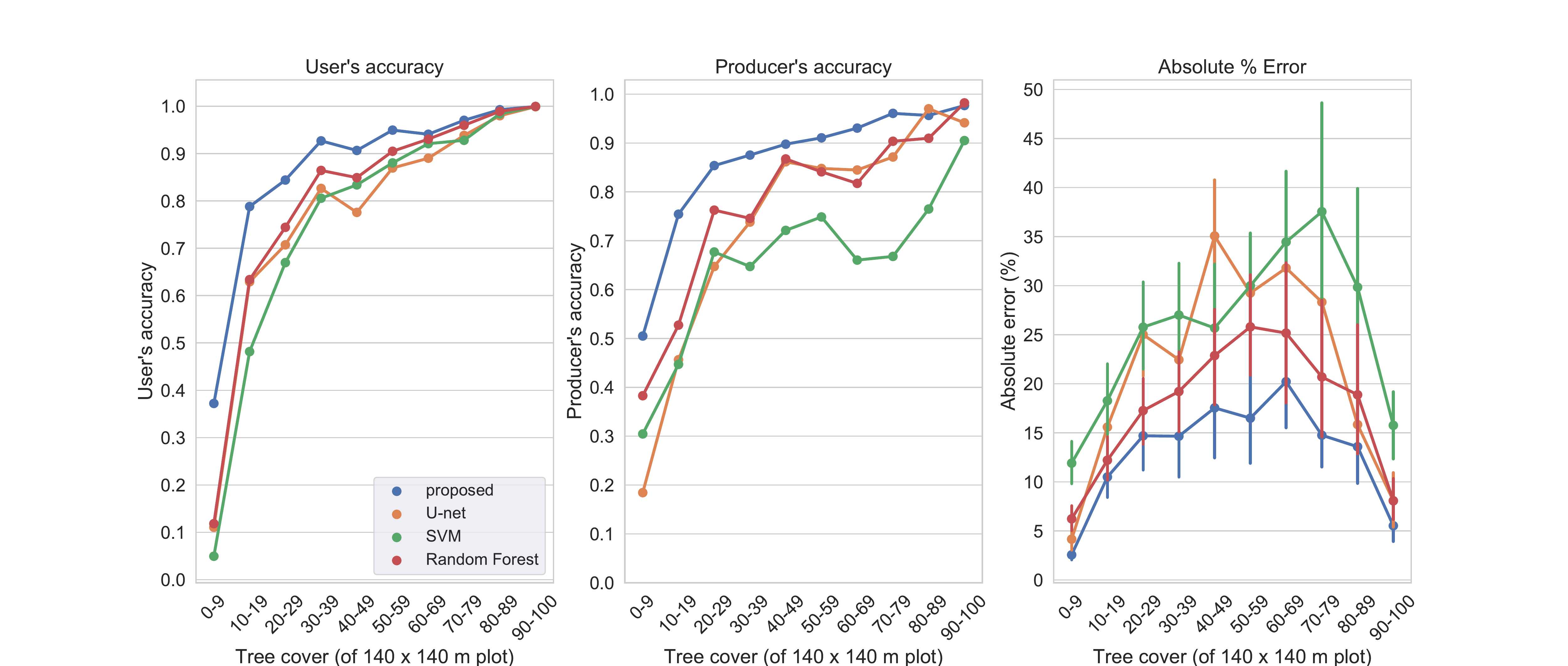}  
\end{tabular}
\end{center}
\caption 
{ \label{fig:treecover}
Performance of proposed model and baseline random forests (RF), support vector machine (SVM), and U-Net CNN measured with user's and producer's accuracy and the absolute percent error for each decile of plot-level tree cover.}
\end{figure*}

\begin{table}[!h] \centering 
  \caption{Per pixel confusion matrix, and user's, producer's, and overall accuracy for background and positive classes.} 
  \label{table:confusionmatrix} 
\begin{tabular}{l|l|c|c|c}
\multicolumn{2}{c}{}&\multicolumn{2}{c}{Label}&\\
\cline{3-4}
\multicolumn{2}{c|}{}&No tree&Tree&\multicolumn{1}{c}{Prod. Acc.}\\
\cline{2-4}
Pred & No tree & $133287$ & $4802$ & $96.5\%$ \\
\cline{2-4}
& Tree & $3599$ & $74304$ & $95.3\%$\\
\cline{2-4}
\multicolumn{1}{c}{} & \multicolumn{1}{c}{Usr. Acc.} & \multicolumn{1}{c}{$97.4\%$} & \multicolumn{    1}{c}{$93.9\%$} & \multicolumn{1}{c}{$OA = 96.1\%$}\\
\end{tabular}
\end{table}

The proposed model is able to identify individual trees and small patches of trees across a variety of tree cover densities, land uses, biomes, terrains, and geographies. The model achieves 94\% user’s accuracy and 95\% producer’s accuracy across the 1,100 test plots, consisting of 215,600 pixels (Table \ref{table:confusionmatrix})). When compared to the baseline random forest, support vector machine, and U-net, the proposed model performs substantially better in low tree cover scenarios, increasing user's and producer's accurracy by 33\% and 24\% over the best performing baseline model (U-net) in plots with below 20\% tree cover (Figure \ref{fig:treecover}; Table \ref{table:results}). For plots with between 20 and 40\% tree cover, the proposed model increases user's and producer's accuracy by 9\% and 13\% over the best baseline model (random forest, Table \ref{table:results}).

\begin{table*}[!h] \centering 
  \caption{Performance of proposed model and baseline models in different geographical and biophysical conditions, measured with user's and producer's accuracy. RF = random forest, SVM = support vector machine.} 
  \label{table:results} 
\begin{tabular}{@{\extracolsep{2pt}} lccccccccccc} 
\\[-1.8ex]\hline 
& \multicolumn{2}{c}{Proposed} & \multicolumn{2}{c}{RF} & \multicolumn{2}{c}{SVM} & \multicolumn{2}{c}{U-net} \\
Parameter & User & Prod & User & Prod & User & Prod & User & Prod & n \\
\hline \\[-1.8ex] 
Region & & & & & &  \\ 
Africa & 0.91 & \textbf{0.95} & 0.91 & 0.85 & 0.79 & 0.79 & 0.93 & 0.78 & 88,984 \\
Asia & \textbf{0.92} & \textbf{0.92}  & 0.90 & 0.88 & 0.86 & 0.84 & 0.90 &0.89 & 24,304 \\
Australia & \textbf{0.93} & \textbf{0.95}  & 0.92 & 0.91 & \textbf{0.95} & 0.62 & 0.93& 0.89 &19,600 \\
Europe & \textbf{0.96} & \textbf{0.96}  & 0.85 & 0.93 & 0.72 & 0.74 & 0.86 & 0.93 & 23,912 \\
N. America & \textbf{0.97} & \textbf{0.97}  & 0.92 & 0.96 & 0.88 & 0.92 & 0.91 & 0.93 & 38,808 \\
S. America & \textbf{0.95} & 0.94  & 0.83& \textbf{0.95} & 0.72 & 0.85 & 0.93 & \textbf{0.95} & 22,148 \\
\hline \\[-1.8ex] 
Cloud (\%) & & & & & & \\ 
0-75 & \textbf{0.93} & \textbf{0.95} & 0.90 & 0.90 & 0.82 & 0.80 & 0.92 & 0.86 & 170,128 \\ 
75+ & \textbf{0.94} & \textbf{0.95} &  0.89 & 0.92 & 0.82 & 0.86 & 0.91 & 0.91 & 45,864 \\ 
\hline \\[-1.8ex] 
Slope (\%) & & & & & & \\
0-10 & \textbf{0.92} & \textbf{0.95} & 0.89 & 0.89 &  0.80 & 0.77 & 0.91 & 0.86 & 176,400 \\ 
10+ & \textbf{0.97} & \textbf{0.95} &  0.91 & \textbf{0.95} & 0.86 & 0.92 & 0.93 & 0.91 & 39,592 \\ 
\hline \\[-1.8ex] 
Canopy (\%) & & & & & & \\
0-20 & \textbf{0.66} & \textbf{0.60} & 0.27 & 0.47 & 0.14 & 0.39 & 0.33 & 0.36 & 115,248 \\ 
20-40 & \textbf{0.88} & \textbf{0.87} & 0.79 & 0.75 & 0.72 & 0.66 & 0.76 & 0.69 & 21,560 \\ 
40-60 & \textbf{0.90} & \textbf{0.93} & 0.88 & 0.85 & 0.86 & 0.74 & 0.83 & 0.85 & 11,956\\ 
60-100 & 0.97 & \textbf{0.99} & \textbf{0.99} & 0.96 & \textbf{0.99} & 0.86 & \textbf{0.99} & 0.93 &  67,228 \\ 
\hline \\[-1.8ex] 
Overall & \textbf{0.94} & \textbf{0.95} & 0.90 & 0.91 & 0.81 & 0.82 & 0.91 & 0.88 & 215,992 \\
\hline \\[-1.8ex] 
\end{tabular} 
\end{table*} 

Across all testing plots, the proposed model improves the mean tree cover error to 7.1\% (6.4 - 7.9\%, 95\% CI) from 10.7\% (8.7 - 11.6\%, random forest). In plots with between 30 and 70\% tree cover, the percent tree cover error is reduced from 22.6\% (20.0 - 25.1\%) to 16.7\% (14.6 - 18.8\%) (Figure \ref{fig:treecover}). The proposed model is not sensitive to cloud cover or steep terrain. In very cloudy test plots (at least 75\% of imagery dates with at least 20\% cloud cover), the proposed model maintains high accuracy, with 95 and 94\% user’s and producer’s accuracy, reducing omission and commission errors against the baseline models by almost half (Table \ref{table:results}). Different methods of temporal aggregation did not considerably change the accuracy metrics for the baseline models (Table \ref{table:aggregation}).

\begin{table*}[!h] \centering 
  \caption{Performance of baseline models with different temporal aggregation methods. Bold numbers indicate the selected aggregation method for the associated model. S.D. = standard deviation.} 
  \label{table:aggregation} 
\begin{tabular}{@{\extracolsep{5pt}} lcccccc} 
\\[-1.8ex]\hline 
\hline \\[-1.8ex] 
& \multicolumn{2}{c}{Random Forest} & \multicolumn{2}{c}{SVM} & \multicolumn{2}{c}{U-net} \\
Aggregation & User & Prod. & User & Prod. & User & Prod. \\ 
\hline \\[-1.8ex] 
Mean  & \textbf{0.90} & \textbf{0.91} & \textbf{0.81} & \textbf{0.82} & 0.91 & 0.88 \\ 
Median  & 0.91 & 0.90 & 0.81 & 0.80 & \textbf{0.92} & \textbf{0.88} \\
Mean + S.D. & 0.89 & 0.90 & 0.82 & 0.81 & 0.90 & 0.90 \\
Median + S.D. & 0.89 & 0.91 & 0.81 & 0.81 & 0.90 & 0.91 \\
Quarterly mean & 0.90 & 0.90 & 0.81 & 0.81 & 0.91 & 0.86 \\
Monthly mean & 0.89 & 0.90 & 0.82 & 0.81 & 0.91 & 0.88 \\ 
\hline \\[-1.8ex] 
\end{tabular} 
\end{table*} 

\begin{figure*}[!htbp]
\begin{center}
\begin{tabular}{c}
\includegraphics[height=4.5cm]{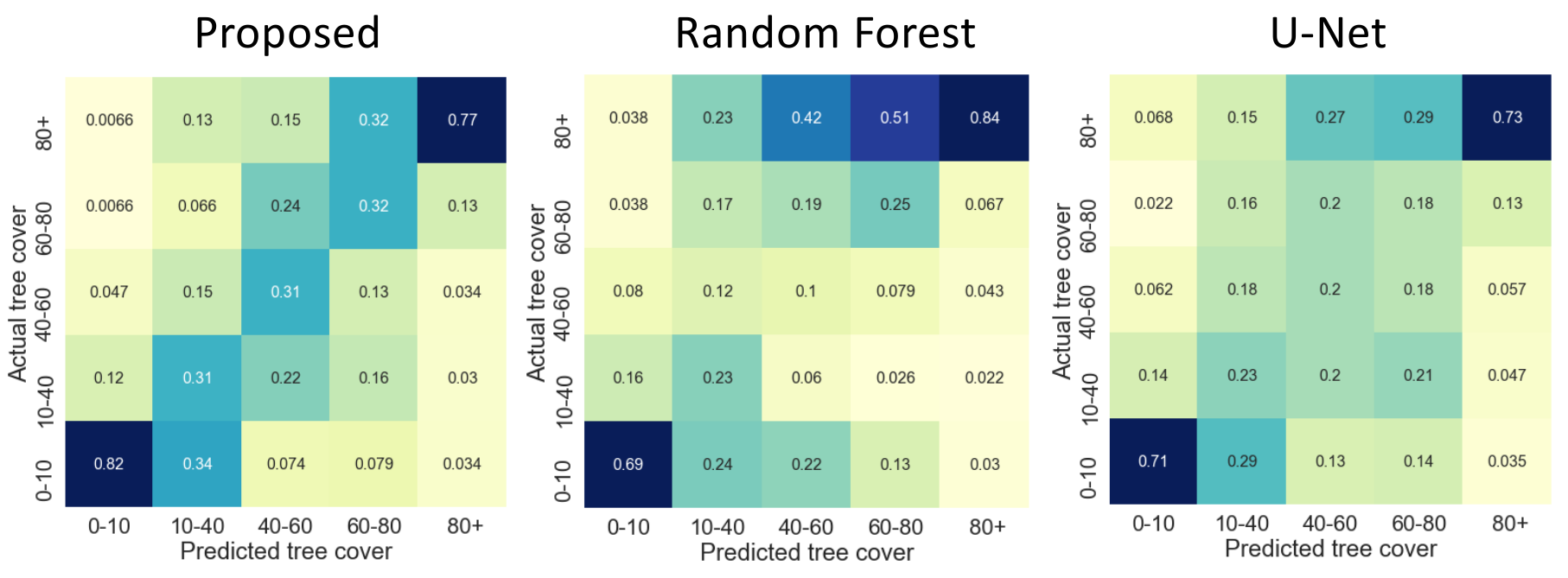}  
\end{tabular}
\end{center}
\caption 
{ \label{fig:bastin}
Confusion matrix of predicted versus labeled tree cover from \citet{bastin2017} for the proposed model, random forest, and U-net.}
\end{figure*}

The predicted tree cover for the 1,000 stratified plot locations from \citet{bastin2017} had an average of 0.85 (0.83-0.86, 95\% CI) Pearson's correlation with the cleaned tree cover labels from \citet{bastin2017}. The Pearson's correlation for the random forest model was 0.76 (0.74-0.79, 95\% CI) and for the U-net was 0.77 (0.74-0.79, 95\% CI). The proposed model classified above and below 10\% tree cover with 92\% and 91\% user's and producer's accuracy, compared to random forest (90\%, 82\%, respectively) and U-net (92\%, 85\%, respectively) (Figure \ref{fig:bastin}).

\subsection{Regional accuracy}

\begin{table*}[!htbp] \centering 
  \caption{User's and producer's accuracy for the selected one million ha regions and 200 ha subregions in Ghana, Tanzania, and Honduras.} 
  \label{table:results-region} 
\begin{tabular}{@{\extracolsep{5pt}} lcccccc} 
\\[-1.8ex]\hline 
\hline \\[-1.8ex] 
& \multicolumn{2}{c}{Proposed model} & \multicolumn{2}{c}{Random Forest} & \multicolumn{2}{c}{U-net} \\
Parameter & User & Prod. & User & Prod. & User & Prod. \\ 
\hline \\[-1.8ex] 
Ghana - region  & \textbf{0.80} & \textbf{0.82} & 0.77 & 0.77 & 0.70 & 0.70 \\ 
Ghana - subregion  & \textbf{0.81} & \textbf{0.76} & 0.78 & 0.75 & 0.75 & 0.76 \\
Tanzania - region & \textbf{0.86} & \textbf{0.84} & 0.75 & 0.78 & 0.74 & 0.84 \\
Tanzania - subregion & \textbf{0.85} & \textbf{0.84} & 0.52 & 0.54 & 0.76 & 0.75 \\
Honduras - region & \textbf{0.99} & \textbf{0.98} & 0.97 & 0.97 & 0.96 & 0.95 \\ 
Honduras - subregion &  \textbf{0.94} & \textbf{0.99} & 0.93 & \textbf{0.99} & 0.92 & 0.97 \\
\hline \\[-1.8ex] 
\end{tabular} 
\end{table*} 

\begin{figure*}[!h]
\begin{center}
\begin{tabular}{c}
\includegraphics[height=12cm]{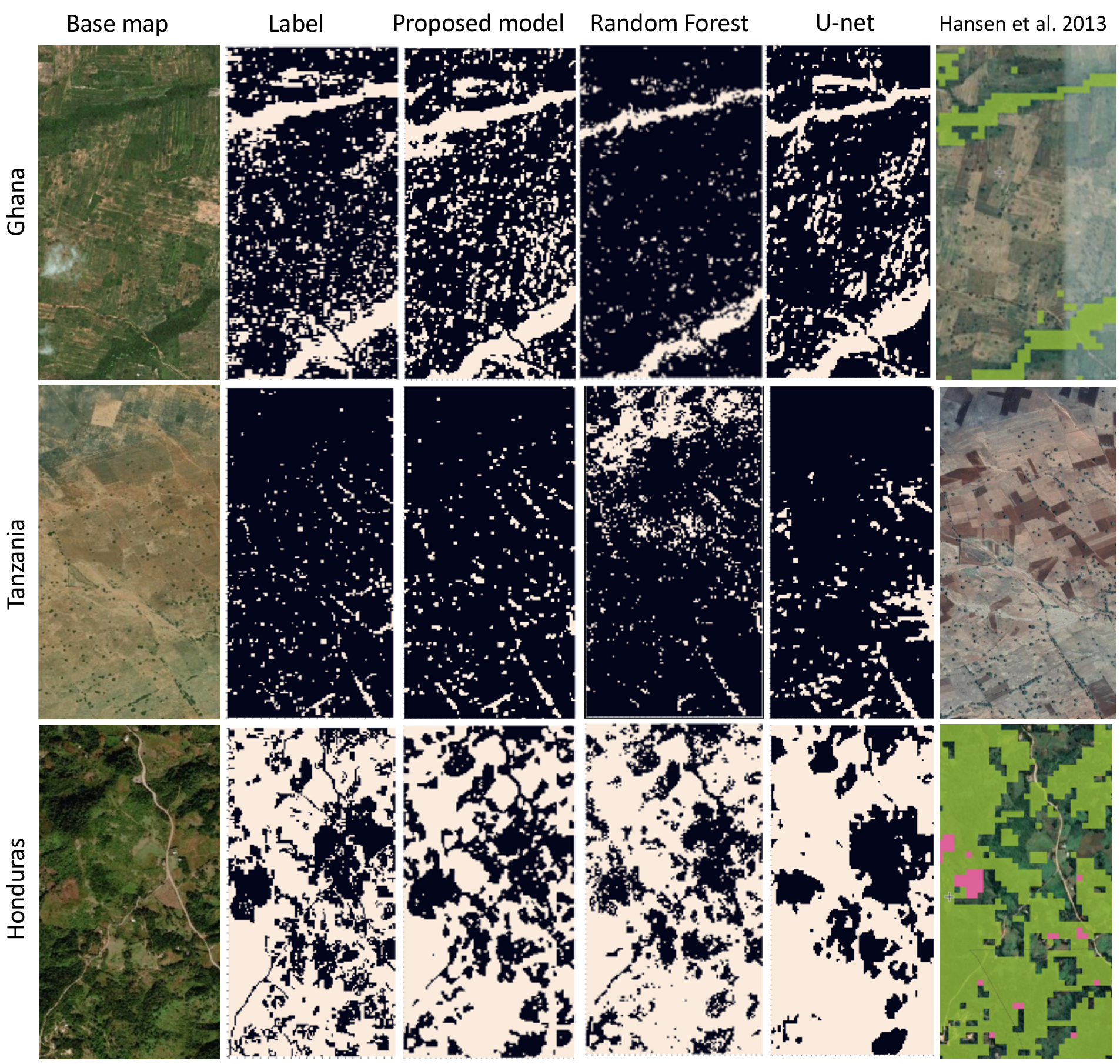}  
\end{tabular}
\end{center}
\caption 
{ \label{fig:large_area}
Predictions of the proposed model and baselines for the 200 hectare subregions in Ghana, Tanzania, and Honduras, along with the labels, based on high-resolution imagery, and the high-resolution base map. Predictions of \citet{hansen2013} are added for reference, where green pixels indicate tree cover, and pink pixels indicate tree cover loss.} 
\end{figure*}

The proposed model achieves an average of 87.5\% user's accuracy and 87.2\% producer's accuracy across the 118,800 labelled pixels in Ghana, Tanzania, and Honduras. This reflects a 7.2\% and 7.3\% increase over the random forests, and a 7\% and 4.2\% increase over the U-net in terms of user's and producer's accuracy (Table \ref{table:results-region}).

In the 200 hectare subregion in Ghana, the proposed model accurately locates both the scattered trees on farm land and the tree corridors (Figure \ref{fig:large_area}). In comparison, the random forest misses the majority of the scattered trees on farm land, while the U-net overpredicts the density of large patches of trees, and completely misses many small patches of trees. Overall, the proposed model reduces relative commission and omission error by 13\% and 22\% versus the random forest, and 33\% and 40\% versus the U-net for the larger region.

The proposed model is able to locate most individual trees in the 200 hectare subregion in Tanzania (Figure \ref{fig:large_area}). The random forest model only detected large patches of trees, and the U-net again over-predicted the density of large patches while missing most of the individual trees. When compared to the baseline random forest model, the proposed model reduces relative commission and omission error by 63\% and 27\% for the larger region, and commission error by 46\% versus the U-net.

\begin{figure}[!h]
\begin{center}
\begin{tabular}{c}
\includegraphics[height=8cm]{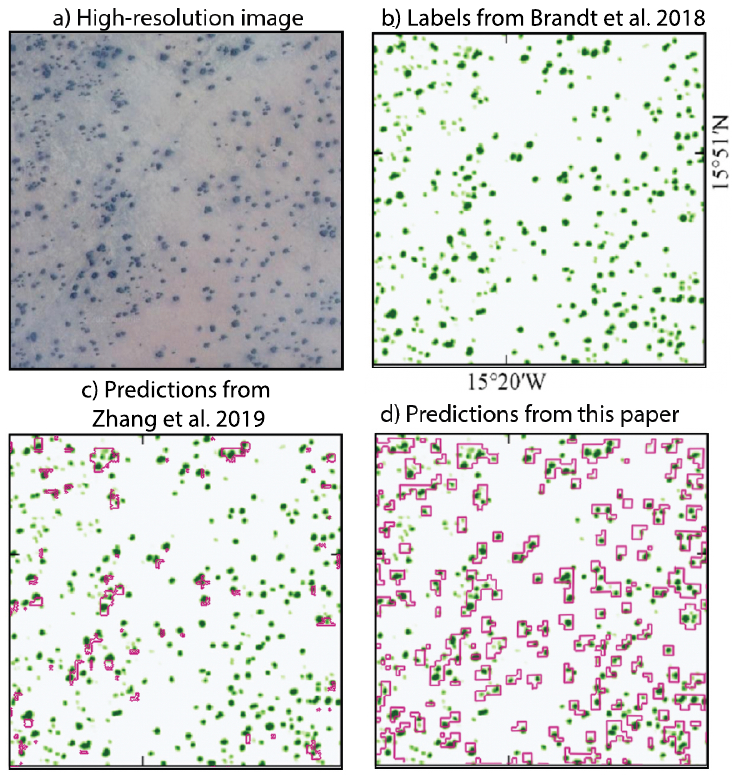}  
\end{tabular}
\end{center}
\caption 
{ \label{fig:zhang}
Performance comparisons with results from \citet{zhang2019} for a 42 hectare region in Senegal. Labels from \citet{brandt2018} are denoted in green and predictions are denoted in magenta. The proposed model identifies 85\% of the trees labeled in \citet{brandt2018}.}
\end{figure} 

In the 200 hectare subregion in Honduras, the proposed model identified most of the small patches of barren land within the surrounding dense tree cover, while also identifying many of the individual trees that on farmland (Figure \ref{fig:large_area}). In comparison, the random forest’s predictions were very noisy around the boundaries between dense and scattered tree cover, and the individual trees on farmland were not identified. The U-net was unable to identify the small patches of barren land, and overestimated tree cover substantially.  When visually compared with the results from \citet{hansen2013}, the proposed model performs much better at identifying scattered trees, while \citet{hansen2013} performs visually similar for trees inside closed-canopy forests (Figure \ref{fig:large_area}). Compared to the predictions of \citet{zhang2019}, the proposed model improves accuracy by 50\% from 35\% to 85\% in a 42 hectare region of Senegal (Figure \ref{fig:zhang}).


\section{Discussion}
\subsection{Implications for global forest monitoring efforts}

Global assessments of tree cover have underestimated tree cover extent in drylands as well as scattered tree cover in urban environments and mosaic landscapes \citep{bastin2017, OTTOSEN2020101947, Milodowski_2017}. This has caused considerable uncertainty in estimating the extent of these forms of tree cover at a global scale. Through human photointerpretation of 500,000 plots with high-resolution satellite imagery, \citet{bastin2017} found that global remote sensing classifiers underestimated dryland forest extent by 40\%. With regard to trees on cropland, \citet{treesonagland} found that more than 40\% of cropland on earth has at least 10\% canopy cover. However, without a globally consistent and efficient method to measure these types of sparse and scattered tree cover, estimates continue to rely either on aggregating data from multiple, small-scale assessments, or on human photointerpretation.

The methodology presented in this paper enables the assessment of sparse and scattered tree cover with medium-resolution satellite imagery. The accuracy of monitoring tree presence in highly heterogeneous areas such as those with sparse and scattered tree cover was improved by as much as 20\% over standard remote sensing classifiers such as random forests and support vector machines. Additionally, the proposed approach reduces omission and commission errors in dense tree cover, high cloud cover, and mountainous regions by nearly half compared to standard remote sensing classifiers. Global data that combines contiguous, closed-canopy forest extent with tree cover data in low tree cover regions will have significant implications for global land use and land change monitoring. 

Many types and drivers of land use change, such as small area deforestation, encroachment, habitat fragmentation, and natural regeneration, occur primarily in highly heterogeneous regions like the urban fringe, forest perimeters, and riparian zones \citep{Tyukavinaeaat2993, woodyencroachment, Pickett331, Rex1990TheFS}. For instance, the majority of woody encroachment into open areas in continental Africa occurs in regions with moderate initial tree cover (between 30 - 60\%) \citep{woodyencroachment}. Furthermore, more than two thirds of deforestation in both the Congo basin and the Amazon between 2000 and 2014 was driven by smallholder clearing \citep{kalamandeen}. Human-driven increases in canopy cover, such as afforestation and reforestation, also tend to be comprised of many small tree planting efforts, often by individual land owners \citep{Holl455, MELO2013395}. These highly heterogeneous regions where important land use changes are occurring are known to have lower classification accuracy for tree cover and land use than larger, homogeneous regions in many remote sensing approaches \citep{Smith2002IMPACTSOP, Milodowski_2017, reddplus}. This limits the applicability of existing global tree cover datasets to monitor small area deforestation, natural regeneration, and landscape restoration.

More than 50 countries have committed to putting more than 170 million hectares of land under restoration by 2030 \citep{fao_wri}. Restoration land use interventions target open-canopy forests and trees outside of forests, including agroforestry, silviculture, and natural regeneration. While global data like that of \citet{hansen2013} have transformed forest monitoring and carbon accounting, there is no comparable global method or data to monitor progress on restoration. Instead, restoration monitoring programs are generally constructed at the national or landscape scale, and rely on field data and human interpretation of satellite imagery \citep{Bey2016CollectEL, afr100}. The methodological improvements for tree identification in areas with sparse and scattered tree cover presented in this paper may allow for increased spatial and temporal resolution of monitoring data for land use management decisions by reducing reliance on photointerpretation and field data.

\subsection{Implications for remote sensing methodologies}
The applications of deep learning methodologies such as CNNs to medium-resolution remote sensing imagery are not well understood. The majority of classification systems rely on random forests or support vector machines rather than CNNs \citep{ma2019}. Indeed, in the present study, the U-net baseline performed similarly to the random forest, despite being more than an order of magnitude more computationally intensive. Many studies have concluded that the U-net significantly outperforms other machine learning classifiers for remote sensing tasks with high-resolution imagery \citep{DBLP:journals/corr/abs-1904-00592, Li2017DeepUNetAD, unet3, unet4, unet5}. The relative underperformance of the U-net with medium-resolution imagery highlights a need for alternative CNN approaches such as the one proposed in this paper.

By explicitly designing the model architecture to mitigate the constraints of medium-resolution data, the proposed model was able to significantly outperform the U-net model and the other machine learning baselines in every decile of tree cover (Figure \ref{fig:treecover}). These design choices included increasing data complexity with temporal data and the convolutional GRU, choosing model building blocks that are explicitly designed for pixel-level accuracy (FPA, csSE), and using a loss function explicitly designed for different tree cover scenarios and coregistration errors (boundary loss and label smoothing). Because the improvements in accuracy are rooted in the temporal CNN architecture, it is likely that the proposed model design could also greatly benefit land use modeling and the identification of objects in satellite imagery such as roofs, power plants, boats, or roads.  

\subsection{Future research}

Despite the numerous improvements presented in this paper, there are several avenues for future research in imagery preprocessing that could further improve accuracy in low tree cover regions. Due to the global nature of this work, the present approach naively fuses Sentinel-1 and Sentinel-2 by simply stacking the Sentinel-1 ground range detected (GRD) product with the Sentinel-2 L2A product. Because of the sub-pixel size of trees in Sentinel imagery, the pixel-level match up between Sentinel-1 and Sentinel-2 is very important for reducing the blurriness of predictions. Many new approaches for fusing Sentinel-1 and Sentinel-2 have recently been proposed, such as probabilistic Bayesian models, affine transformations, and CNN-based approaches \citep{plsa, fuse}. While these approaches are often limited to small geographies and their benefit to global-scale preprocessing is not yet clear, increasing the pixel-level accuracy of Sentinel-1 and 2 fusion would likely improve the identification of small tree patches.

Because the proposed model uses multitemporal imagery, the interpolation of cloud and cloud shadow pixels is very important to accurately reconstruct imagery, especially when the gap between cloud-free image acquisitions is substantial. Although the proposed model in this paper performs much better than the baseline models in high cloud cover scenarios, there is room to improve the interpolation of cloud and cloud shadow when constructing time series imagery. The presented methodology uses the Whittaker smoother, which \citet{whittakermodis} found to be better at interpolating time series gaps in MODIS imagery in comparison to temporal smoothing and gap filling, Gaussian functions, low pass filters, and many other per-pixel smoothing functions. In contrast to these per-pixel approaches, a number of spatio-temporal methods for gap filling have been proposed. Such methods include conditional generative adversarial networks \citep{gan1}, convolutional neural networks \citep{ZHANG2020148}, patch-based cloning \citep{patchclone}, k-nearest neighbors \citep{knn}, and spatio-temporal Markov random fields \citep{CHENG201454}.

Finally, future research should identify the benefits of the proposed model for change detection, such as identifying small-scale deforestation as well as forest and landscape restoration activities. While the present paper demonstrates comparative benefits over commonly used remote sensing classifiers for tree identification in heterogeneous landscapes, the potential benefits for change detection are left to future work. In a similar manner, future research should detail how the methodology presented in this paper can be integrated into national and subnational environmental monitoring programs. One potential avenue is through human-in-the-loop artificial intelligence, where locally generated data is used to improve the local accuracy of globally developed models \citep{logar2020pulsesatellite}. Because neural networks allow for online learning, they can easily be 'fine-tuned' with new data, while models such as random forests and support vector machines cannot \citep{Zhou_2017_CVPR}. This allows for the combination of participatory mapping and crowd sourcing, where local stakeholders label high-resolution imagery with the power of neural network based remote sensing classifiers.

\bibliographystyle{tfcad}
\bibliography{report}

\end{document}